\documentclass[pmlr]{jmlr}% new name PMLR (Proceedings of Machine Learning)

\usepackage{longtable}% for long tables

\usepackage{booktabs}
\usepackage{import}
\usepackage{graphicx}     
\usepackage{algorithm}
\usepackage{algorithmicx}
\usepackage{algpseudocode}
\usepackage{bm}

\usepackage{xcolor}
\usepackage{extarrows}
\usepackage{multirow}
\usepackage[load-configurations=version-1]{siunitx} % newer version
\theorembodyfont{\upshape}
\theoremheaderfont{\scshape}
\theorempostheader{:}
\theoremsep{\newline}

\jmlrvolume{1}
\jmlryear{2019}
\jmlrworkshop{NeurIPS2019 Disentanglement Challenge}

\usepackage{wrapfig}
\usepackage[T1]{fontenc} % fix Command \DJ unavailable in encoding OT1

\title{Disentanglement Challenge: From Regularization to Reconstruction}

  \author{\Name{Jie Qiao\nametag{\thanks{equal contribution}}} \Email{qiaojie.chn@gmail.com}\\
  	\Name{Zijian Li\nametag{$^*$}} \Email{leizigin@gmail.com}\\
  	\Name{Boyan Xu\nametag{$^*$}} \Email{hpakyim@gmail.com}\\
	\Name{Ruichu Cai} \Email{cairuichu@gdut.edu.cn}\\
	\addr School of Computer Science, Guangdong University of Technology, China \AND
	\Name{Kun Zhang} \Email{kunz1@cmu.edu}\\
	\addr Department of philosophy, Carnegie Mellon University }

\begin{document}

\maketitle

\begin{abstract}

The challenge of learning disentangled representation has recently attracted much attention and boils down to a competition\footnote{\url{https://www.aicrowd.com/challenges/neurips-2019-disentanglement-challenge}\label{web}} using a new real world disentanglement dataset \citep{gondal2019transfer}. Various methods based on variational auto-encoder have been proposed to solve this problem, by enforcing the independence between the representation and modifying the regularization term in the variational lower bound. However recent work by \cite{locatello2018challenging} has demonstrated that the proposed methods are heavily influenced by randomness and the choice of the hyper-parameter. In this work, instead of designing a new regularization term, we adopt the FactorVAE but improve the reconstruction performance and increase the capacity of network and the training step. The strategy turns out to be very effective and achieve the 1st place in the challenge.

\end{abstract}
\begin{keywords}
disentangled representation, unsupervised learning
\end{keywords}

\section{Introduction}
\label{sec:intro}

The great success of unsupervised learning heavily depends on the representation of the feature in the real-world. It is widely believed that the real-world data is generated by a few explanatory factors which are distributed, invariant, and disentangled \citep{bengio2013representation}. The challenge of learning disentangled representation boils down into a competition\textsuperscript{\ref{web}} using a new real world disentanglement dataset \citep{gondal2019transfer} to build the best disentangled model.

The key idea in disentangled representation is that the perfect representation should be a one-to-one mapping to the ground truth disentangled factor. Thus, if one factor changed and other factors fixed, then the representation of the fixed factor should be fixed accordingly, while others' representation changed. As a result, it is essential to find representations that (i) are independent of each other, and (ii) align to the ground truth factor.

Recent line of works in disentanglement representation learning are commonly focused on enforcing the independence of the representation by modifying the regulation term in the variational lower bound of Variational Autoencoders (VAE) \citep{kingma2013auto}, including $\beta$-VAE \citep{higgins2017beta}, AnnealedVAE \citep{burgess2018understanding}, $\beta$-TCVAE \citep{chen2018isolating}, DIP-VAE \citep{kumar2018variational} and FactorVAE \citep{kim2018disentangling}. See Appendix \ref{apd:first} for more details of these model.

To evaluate the performance of disentanglement, several metrics have been proposed, including the FactorVAE metric \citep{kim2018disentangling}, Mutual Information Gap (MIG) \citep{chen2018isolating}, DCI metric \citep{eastwood2018framework}, IRS metric \citep{suter2019robustly}, and SAP score \citep{kumar2018variational}.

\begin{figure*}
	\centering
	\includegraphics[width=\columnwidth]{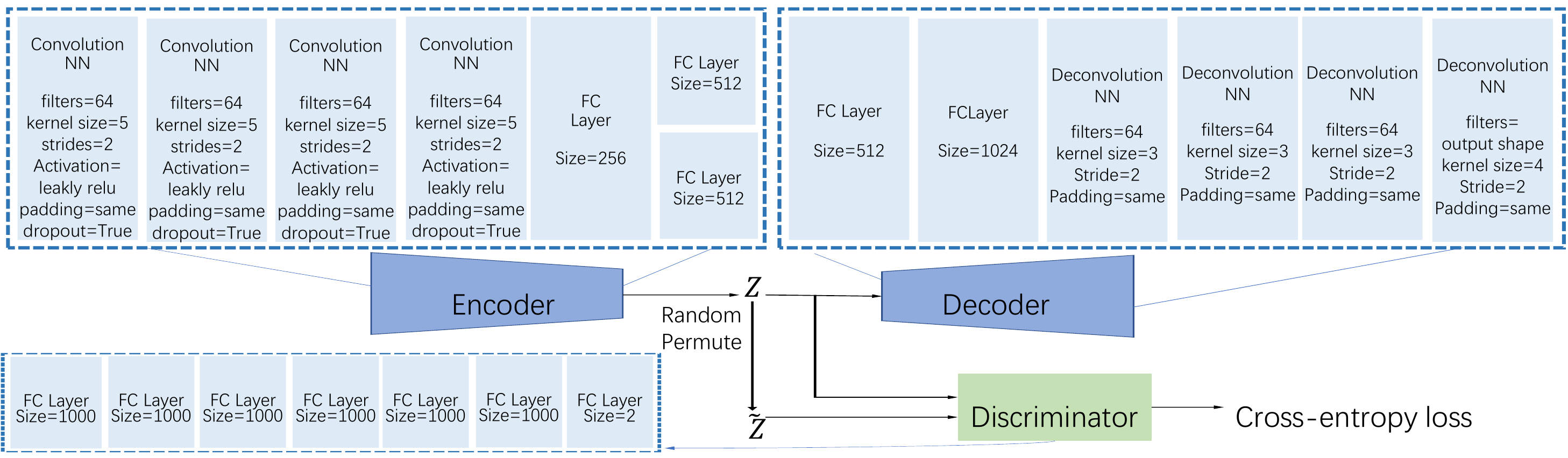}
	\vspace{-1cm}
	\caption{The architecture of FactorVAE, in which FC layer denote the full connection layer}
	\label{fig:model}
	\vspace{-0.9cm}
\end{figure*}

However, one of our findings is that these methods are heavily influenced by randomness and the choice of the hyper-parameter. This phenomenon was also discovered by \cite{locatello2018challenging}. Therefore, rather than designing a new regularization term, we simply use FactorVAE but at the same time improve the reconstruction performance. 
%Furthermore, it is necessary to run experiments with different random seeds. 
We believe that, the better the reconstruction, the better the alignment of the ground-truth factors. Therefore, the more capacity of the encoder and decoder network, the better the result would be. Furthermore, after increasing the capacity, we also try to increase the training step which also shows a significant improvement of evaluation metrics. The final architecture of FactorVAE is given in Figure \ref{fig:model}. Note that, this report contain the results from both stage 1 and stage 2 in the competition.

Overall, our contribution can be summarized as follow: (1) we found that the performance of the reconstruction is also essential for learning disentangled representation, and (2) we achieve state-of-the-art performance in the competition.

\section{Experiments Design}

In this section, we explore the effectiveness of different disentanglement learning models and the performance of the reconstruction for disentangle learning. We first employ different kinds of variational autoencoder including \textit{BottleneckVAE}, \textit{AnneledVAE}, \textit{DIPVAE}, \textit{BetaTCVAE}, and \textit{BetaVAE} with $30000$ training step. Second, we want to know whether the capacity plays an important role in disentanglement. The hypothesis is that the larger the capacity, the better reconstruction can be obtained, which further reinforces the disentanglement. In detail, we control the number of latent variables.

\section{Experiments Results}

In this section, we present our experiment result in stage 1 and stage 2 of the competition. We first present the performance of different kinds of VAEs in stage 1, which is given in Table \ref{tab:result1}. It shows that FactorVAE achieves the best result when the training step is 30000. In the following experiment, we choose FactorVAE as the base model.

\begin{table}[htbp]
	\centering
	\footnotesize
	\caption{Variational autoencoder with 30000 training steps in Stage 1.}\label{tab:result1}
	\begin{tabular}{l|c|c|c|c|c}
		\toprule
		% after \\: \hline or \cline{col1-col2} \cline{col3-col4} ...
		VAE variation & FactorVAE & sap score & dci   & irs   & mig   \\
		\midrule
		BottleneckVAE& 0.453     & 0.0395    & 0.107  & 0.547 & 0.0589\\
		AnneledVAE   & 0.3586    & 0.0069    & 0.1153 & 0.5122& 0.0237\\
		DIPVAE        & 0.265     & 0.005     & 0.021  & 0.265 & 0.490 \\
		BetaTCVAE     & 0.342     & 0.026     & 0.093  & 0.342 & 0.981\\
		BetaVAE      & 0.3586    & 0.0069    & 0.1153 & 0.5122 & 0.0237\\
		FactorVAE    & $\bm{0.449}$&$\bm{0.0596}$&$\bm{0.1385}$&$\bm{0.5976}$&$\bm{0.0589}$\\
		\bottomrule
	\end{tabular}
\end{table}

Furthermore, we find that (i) the activation function at each layer and that (ii) the size of latent variables are propitious to the disentanglement performance. Therefore, Leaky ReLU and the latent size of 256 are selected in stage 1. Then, as shown in Table \ref{tab:result3}, we increase the step size and we find that the best result was achieved at 1000k training steps. The experiment in this part may not be sufficient, but it still suggests the fact that the larger the capacity is, the better the disentanglement performance. Since we increase the capacity of the model, it is reasonable to also increase the training steps at the same time. In stage 2, as shown in Table \ref{tab:stage2}, using sufficient large training step ($\ge 800k$), we investigate the effectiveness of the number of latent variables. This experiment suggests that the FactorVAE and the DCI metric are positive as the latent variables increase, while the other metrics decrease. The best result in the ranking is marked as bold, which suggests that we should choose an appropriate number of latent variables.

\begin{table}[tbh]
	\vspace{-0.5cm}
	\centering
	\footnotesize
	\caption{FactorVAE with different training step in Stage 1.}\label{tab:result3}
	\begin{tabular}{p{4cm}|c|c|c|c|c}
		\toprule
		training step  & FactorVAE & sap score & dci      & irs      & mig   \\ 
		\midrule
		FactorVAE (30k)                            & 0.449    &${0.0596}$  &${0.1385}$&${0.5976}$&${0.0589}$\\
		FactorVAE (500k)                           & 0.432   &${0.0743}$  &${0.4395}$&${0.6041}$&${0.0739}$\\
		FactorVAE (1000k)                           & $\bm{0.4844}$ &$\bm{0.155}$  &$\bm{0.523}$&$\bm{0.6205}$&$\bm{0.3887}$ \\ % heyman version
		\bottomrule
	\end{tabular}
	\vspace{-0.3cm}
\end{table}

\begin{table}[htbp]
	\centering
	\footnotesize
	\vspace{-0.5cm}
	\caption{Sensitivity analysis for the number of latent variables in Stage 2}
	\begin{tabular}{l|c|c|c|c|c}
		\toprule
		\multicolumn{1}{l|}{Num. of Latent} & \multicolumn{1}{l|}{FactorVAE} & \multicolumn{1}{l|}{sap score} & \multicolumn{1}{l|}{dci} & \multicolumn{1}{l|}{irs} & \multicolumn{1}{l}{mig} \\
		\midrule
		256   & 0.4458 & 0.1748 & 0.5785 & 0.5553 & 0.4166 \\
		512   & \textbf{0.5018} & \textbf{0.1545} & \textbf{0.5457} & \textbf{0.6824} & \textbf{0.3739} \\
		768   & 0.4526 & 0.0942 & 0.5154 & 0.4638 & 0.2793 \\
		1024  & 0.4708 & 0.0955 & 0.542 & 0.4867 & 0.2781 \\
		1536  & 0.5202 & 0.008 & 0.5932 & 0.5071 & 0.0058 \\
		2048  & 0.5374 & 0.0025 & 0.6351 & 0.5218 & 0.0062 \\
		3072  & 0.5426 & 0.0053 & 0.6677 & 0.5067 & 0.0117 \\
		\bottomrule
	\end{tabular}%
	\vspace{-0.9cm}
	\label{tab:stage2}
\end{table}

\section{Conclusion}

In this work, we conducted an empirical study on disentangled learning. We first conduct several experiments with different disentangle learning methods and select the FactorVAE as the base model; and second we improve the performance of the reconstruction, by increasing the capacity of the model and the training step. Finally, our results appear to be competitive.

\bibliography{jmlr-sample}

\begin{thebibliography}{11}
\providecommand{\natexlab}[1]{#1}
\providecommand{\url}[1]{\texttt{#1}}
\expandafter\ifx\csname urlstyle\endcsname\relax
  \providecommand{\doi}[1]{doi: #1}\else
  \providecommand{\doi}{doi: \begingroup \urlstyle{rm}\Url}\fi

\bibitem[Bengio et~al.(2013)Bengio, Courville, and
  Vincent]{bengio2013representation}
Yoshua Bengio, Aaron Courville, and Pascal Vincent.
\newblock Representation learning: A review and new perspectives.
\newblock \emph{IEEE transactions on pattern analysis and machine
  intelligence}, 35\penalty0 (8):\penalty0 1798--1828, 2013.

\bibitem[Burgess et~al.(2018)Burgess, Higgins, Pal, Matthey, Watters,
  Desjardins, and Lerchner]{burgess2018understanding}
Christopher~P Burgess, Irina Higgins, Arka Pal, Loic Matthey, Nick Watters,
  Guillaume Desjardins, and Alexander Lerchner.
\newblock Understanding disentangling in beta-vae.
\newblock \emph{arXiv preprint arXiv:1804.03599}, 2018.

\bibitem[Chen et~al.(2018)Chen, Li, Grosse, and Duvenaud]{chen2018isolating}
Tian~Qi Chen, Xuechen Li, Roger~B Grosse, and David~K Duvenaud.
\newblock Isolating sources of disentanglement in variational autoencoders.
\newblock In \emph{Advances in Neural Information Processing Systems}, pages
  2610--2620, 2018.

\bibitem[Eastwood and Williams(2018)]{eastwood2018framework}
Cian Eastwood and Christopher~KI Williams.
\newblock A framework for the quantitative evaluation of disentangled
  representations.
\newblock 2018.

\bibitem[Gondal et~al.(2019)Gondal, W{\"u}thrich, Miladinovi{\'c}, Locatello,
  Breidt, Volchkov, Akpo, Bachem, Sch{\"o}lkopf, and Bauer]{gondal2019transfer}
Muhammad~Waleed Gondal, Manuel W{\"u}thrich, {\DJ}or{\dj}e Miladinovi{\'c},
  Francesco Locatello, Martin Breidt, Valentin Volchkov, Joel Akpo, Olivier
  Bachem, Bernhard Sch{\"o}lkopf, and Stefan Bauer.
\newblock On the transfer of inductive bias from simulation to the real world:
  a new disentanglement dataset.
\newblock \emph{arXiv preprint arXiv:1906.03292}, 2019.

\bibitem[Higgins et~al.(2017)Higgins, Matthey, Pal, Burgess, Glorot, Botvinick,
  Mohamed, and Lerchner]{higgins2017beta}
Irina Higgins, Loic Matthey, Arka Pal, Christopher Burgess, Xavier Glorot,
  Matthew Botvinick, Shakir Mohamed, and Alexander Lerchner.
\newblock beta-vae: Learning basic visual concepts with a constrained
  variational framework.
\newblock \emph{ICLR}, 2\penalty0 (5):\penalty0 6, 2017.

\bibitem[Kim and Mnih(2018)]{kim2018disentangling}
Hyunjik Kim and Andriy Mnih.
\newblock Disentangling by factorising.
\newblock In \emph{International Conference on Machine Learning}, pages
  2654--2663, 2018.

\bibitem[Kingma and Welling(2013)]{kingma2013auto}
Diederik~P Kingma and Max Welling.
\newblock Auto-encoding variational bayes.
\newblock \emph{arXiv preprint arXiv:1312.6114}, 2013.

\bibitem[Kumar et~al.(2018)Kumar, Sattigeri, and
  Balakrishnan]{kumar2018variational}
Abhishek Kumar, Prasanna Sattigeri, and Avinash Balakrishnan.
\newblock Variational inference of disentangled latent concepts from unlabeled
  observations.
\newblock In \emph{International Conference on Learning Representations}, 2018.
\newblock URL \url{https://openreview.net/forum?id=H1kG7GZAW}.

\bibitem[Locatello et~al.(2018)Locatello, Bauer, Lucic, Gelly, Sch{\"o}lkopf,
  and Bachem]{locatello2018challenging}
Francesco Locatello, Stefan Bauer, Mario Lucic, Sylvain Gelly, Bernhard
  Sch{\"o}lkopf, and Olivier Bachem.
\newblock Challenging common assumptions in the unsupervised learning of
  disentangled representations.
\newblock \emph{arXiv preprint arXiv:1811.12359}, 2018.

\bibitem[Suter et~al.(2019)Suter, Miladinovic, Sch{\"o}lkopf, and
  Bauer]{suter2019robustly}
Raphael Suter, Djordje Miladinovic, Bernhard Sch{\"o}lkopf, and Stefan Bauer.
\newblock Robustly disentangled causal mechanisms: Validating deep
  representations for interventional robustness.
\newblock In \emph{International Conference on Machine Learning}, pages
  6056--6065, 2019.

\end{thebibliography}

\appendix

\section{Related works}\label{apd:first}

In this section, we are going to summarize the state-of-the-art unsupervised disentanglement learning methods. Most of works are developed based on the Variational Auto-encoder (VAE) \citep{kingma2013auto}, a generative model that maximize the following evidence lower bound to approximate the intractable distribution $p_{\theta}(\mathbf{x} | \mathbf{z})$ using $q_{\phi}(\mathbf{z} | \mathbf{x})$, 
\begin{equation}
\label{eq:elbo}
\max_{\phi ,\theta } \ \ \underbrace{\mathbb{E}_{p(\mathbf{x} )}[\mathbb{E}_{q_{\phi } (\mathbf{z} |\mathbf{x} )}[\log p_{\theta } (\mathbf{x} |\mathbf{z} )]}_{Reconstruction\ Loss} -\underbrace{D_{\mathrm{KL}}( q_{\phi } (\mathbf{z} |\mathbf{x} )\| p(\mathbf{z} ))]}_{Regularization} ,
\end{equation}
where $q_{\phi}(\mathbf{z} | \mathbf{x})$ denote \textit{Encoder} with parameter $\mathbf{\phi}$ and $p_{\theta}(\mathbf{x} | \mathbf{z})$ denote \textit{Decoder} with parameter $\mathbf{\theta}$. 

As shown in Table \ref{tab:related}, all the lower bound of variant VAEs can be described as $Reconstruction\ Loss + Regularization$ where all the Regularization term and the hyper-parameters are given in this table.

\begin{table}[htbp]
	\centering
	\tiny
	\caption{Summary of variant unsupervised disentanglement learning methods}
	\begin{tabular}{l|l|l}
		\toprule
		Model & Regularization & Hyper-Parameters \\
		\midrule
		$\beta$-VAE & $
		\beta\left.D_{\mathrm{KL}}\left(q_{\phi}(\mathbf{z} | \mathbf{x}) \| p(\mathbf{z})\right)\right]
		$ & $\beta$ \\
		AnnealedVAE &  $\left.\gamma\left|D_{\mathrm{KL}}\left(q_{\phi}(\mathbf{z} | \mathbf{x}) \| p(\mathbf{z})\right)-C\right|\right]$ & $\gamma, C$ \\
		FactorVAE & $\left.D_{\mathrm{KL}}\left(q_{\phi}(\mathbf{z} | \mathbf{x}) \| p(\mathbf{z})\right)\right]+\gamma D_{\mathrm{KL}}\left(q(\mathbf{z}) \| \prod_{j=1}^{d} q\left(\mathbf{z}_{j}\right)\right)$ & $\gamma$ \\
		DIP-VAE-I & $\left.D_{\mathrm{KL}}\left(q_{\phi}(\mathbf{z} | \mathbf{x}) \| p(\mathbf{z})\right)\right]+\lambda_{o d} \sum_{i \neq j}\left[\operatorname{Cov}_{p(\mathbf{x})}\left[\mu_{\phi}(\mathbf{x})\right]\right]_{i j}^{2}+\lambda_{d} \sum_{i}\left(\left[\operatorname{Cov}_{p(\mathbf{x})}\left[\mu_{\phi}(\mathbf{x})\right]\right]_{i i}-1\right)^{2}$ & $\lambda_{od}, \lambda_{d}$ \\
		DIP-VAE-II & $\left.D_{\mathrm{KL}}\left(q_{\phi}(\mathbf{z} | \mathbf{x}) \| p(\mathbf{z})\right)\right]+\lambda_{o d} \sum_{i \neq j}\left[\operatorname{Cov}_{q_{\phi}}[\mathbf{z}]\right]_{i j}^{2}+\lambda_{d} \sum_{i}\left(\left[\operatorname{Cov}_{q_{\phi}}[\mathbf{z}]\right]_{i i}-1\right)^{2}$ & $\lambda_{od}, \lambda_{d}$ \\
		\bottomrule
	\end{tabular}%
	\label{tab:related}%
\end{table}%

\end{document}